\documentclass[journal]{IEEEtran}
\usepackage{amsmath,amsfonts}
\usepackage{algorithmic}
\usepackage{array}
\usepackage[caption=false,font=normalsize,labelfont=sf,textfont=sf]{subfig}
\usepackage{textcomp}
\usepackage{stfloats}
\usepackage{url}
\usepackage{verbatim}
\usepackage{graphicx}
\usepackage{svg}
\usepackage{amsmath}
\usepackage{amssymb}
\usepackage{multirow}
\usepackage{booktabs} 
\usepackage{xcolor}
\usepackage{array} 
\usepackage{adjustbox} 
\usepackage{float}
\usepackage{subfig}  
\usepackage{stfloats}
\usepackage{etoolbox}
\usepackage{orcidlink}
\AtEndEnvironment{table}{\vspace{-10pt}}
\AtEndEnvironment{table*}{\vspace{-10pt}}
\AtEndEnvironment{figure}{\vspace{-10pt}}
\AtEndEnvironment{figure*}{\vspace{-10pt}}
\newcolumntype{C}{>{\centering\arraybackslash}m{2.1cm}}

\def\BibTeX{{\rm B\kern-.05em{\sc i\kern-.025em b}\kern-.08em
    T\kern-.1667em\lower.7ex\hbox{E}\kern-.125emX}}
\usepackage{balance}
\usepackage{hyperref}
\hypersetup{
    colorlinks=true,
    linkcolor=black,
    citecolor=black,
    filecolor=black,
    urlcolor=black
}
\begin{document}
\title{SMFusion: Semantic-Preserving Fusion of Multimodal Medical Images for Enhanced Clinical Diagnosis}
\author{Haozhe Xiang~\orcidlink{0009-0008-5810-5426}, Han Zhang~\orcidlink{0000-0001-8498-3451}, Yu Cheng, Xiongwen Quan, Wanwan Huang~\orcidlink{0000-0003-4578-5078}
\thanks{This paper is supported by National Natural Science Foundation of China 62403194.(\textit{Corresponding author: Wanwan Huang.})

Haozhe Xiang and Wanwan Huang are with the College of Information and Intelligence, Hunan Agricultural University, Changsha 410128, China (e-mail: haozhexiang@163.com; hnnydx1993@163.com).

Han Zhang, Yu Cheng and Xiongwen Quan are with the College of Artificial Intelligence, Nankai University, Tianjin 300071, China (e-mail: zhanghan@nankai.edu.cn; yucheng@mail.nankai.edu.cn; quanxw@nankai.edu.cn)}}


\maketitle

\begin{abstract}
Multimodal medical image fusion plays a crucial role in medical diagnosis by integrating complementary information from different modalities to enhance image readability and clinical applicability. However, existing methods mainly follow computer vision standards for feature extraction and fusion strategy formulation, overlooking the rich semantic information inherent in medical images. To address this limitation, we propose a novel semantic-guided medical image fusion approach that, for the first time, incorporates medical prior knowledge into the fusion process. Specifically, we construct a publicly available multimodal medical image-text dataset, upon which text descriptions generated by BiomedGPT are encoded and semantically aligned with image features in a high-dimensional space via a semantic interaction alignment module. During this process, a cross attention based linear transformation automatically maps the relationship between textual and visual features to facilitate comprehensive learning. The aligned features are then embedded into a text-injection module for further feature-level fusion. Unlike traditional methods, we further generate diagnostic reports from the fused images to assess the preservation of medical information. Additionally, we design a medical semantic loss function to enhance the retention of textual cues from the source images. Experimental results on  test datasets demonstrate that the proposed method achieves superior performance in both qualitative and quantitative evaluations while preserving more critical medical information.
\end{abstract}

\section{Introduction}
\label{sec:intro}
\IEEEPARstart{M}{edical} imaging plays a crucial role in modern healthcare, facilitating disease diagnosis, treatment planning, and surgical navigation. Due to differences in imaging principles, various imaging modalities provide complementary information. For instance, computed tomography (CT) offers high-resolution details of bones and dense tissues, whereas magnetic resonance imaging (MRI) provides high-contrast visualization of muscles, nerves, and soft tissues. Functional imaging techniques such as positron emission tomography (PET) and single-photon emission computed tomography (SPECT) reveal metabolic activities and blood flow distribution, making them valuable for tumor detection and neurological disease diagnosis\cite{fu2020multimodal}. However, each imaging modality has inherent limitations. In clinical practice, the information captured in a single imaging modality is often insufficient for comprehensive diagnosis and treatment planning\cite{yin2018medical}. 
\begin{figure}[htbp]
    \centering
    \includegraphics[width=\linewidth,trim=18 18 18 18, clip]{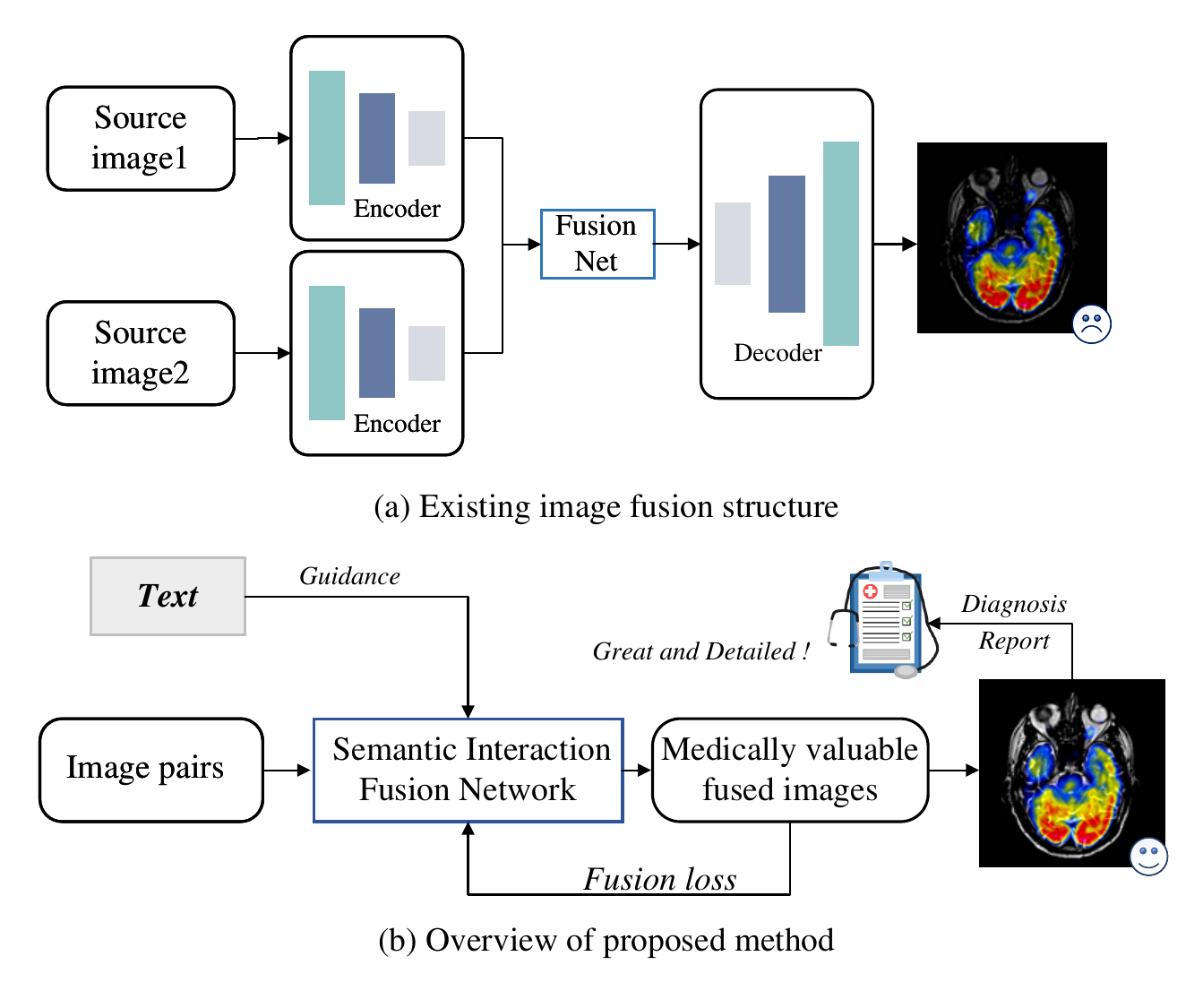} 
    \caption{\small Comparison of Existing Medical Image Fusion Methods and Our Proposed Approach. (a) Existing fusion methods: These methods are primarily based on deep learning networks and integrate information from multiple source images through unsupervised learning strategies. (b) Proposed Semantic-Guided Image Fusion: Our approach integrates medical prior knowledge to explicitly guide the fusion process, establishing an interactive framework that ensures high-quality fusion results.}
    \label{fig:liuchentu1}
\end{figure}
Each modality captures distinct but complementary features, which requires the integration of multiple sources to achieve a more comprehensive and accurate representation of the patient’s condition. Medical image fusion addresses this need by combining information from various imaging modalities into a single, more informative image. By leveraging fusion techniques, clinicians can access structural, functional, and even molecular information simultaneously, facilitating improved diagnosis, treatment planning, and intraoperative guidance. The fused image must preserve critical anatomical details while retaining complementary modality-specific features, ensuring that no clinically relevant information is lost.

Multimodal Medical Image Fusion (MMIF) techniques can be broadly categorized into traditional approaches and deep learning-based methods. Traditional techniques, including spatial domain methods, transform domain approaches, sparse representation models, and hybrid strategies, have been extensively applied in multimodal image integration. Spatial domain methods, such as pixel-level weighted averaging\cite{nunez1999multiresolution} and Principal Component Analysis (PCA)\cite{mackiewicz1993principal}, offer computational simplicity but often result in blurred boundaries and the loss of essential anatomical 
\begin{figure}[htbp]
    \centering
    \makebox[\linewidth]{\hspace{-0.01\linewidth}\includegraphics[width=\linewidth,trim=18 18 18 18, clip]{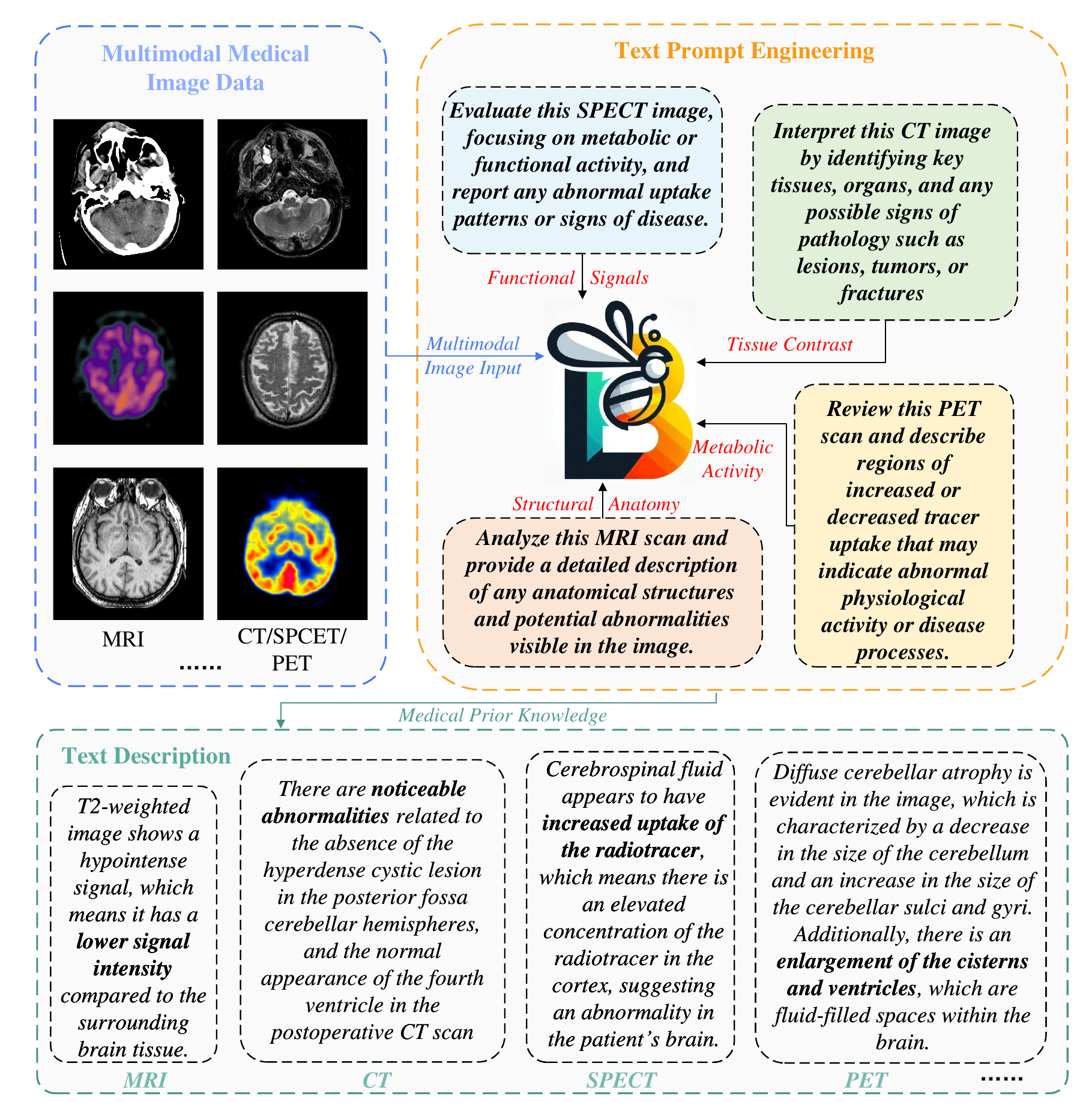}}
    \caption{Schematic illustration of multimodal medical image-text dataset construction. For each imaging modality, tailored text prompts are provided to guide BiomedGPT in generating corresponding descriptions.}
    \label{fig:liuchentu2}
\end{figure}
details due to their limited capacity to model spatial dependencies.Transform domain approaches, including Wavelet Transform (WT)\cite{chai2017image}, curvelet transform \cite{ma2010curvelet}, and Nonsubsampled Shearlet Transform (NSST)\cite{yang2014image}, improve fusion by enabling multi-scale and multi-directional analysis. However, their reliance on manually designed decomposition frameworks and heuristic fusion rules limits their flexibility and robustness across different clinical contexts. Sparse representation methods\cite{li2023sparse} aim to enhance detail preservation and noise robustness by learning sparse coefficients over overcomplete dictionaries. Despite their effectiveness, these methods typically involve high computational costs and are sensitive to the choice of dictionary. Hybrid strategies\cite{wang2020multi} combine multiple paradigms to capitalize on their respective advantages, but this often introduces additional complexity and parameter tuning challenges, reducing practical applicability in real-time scenarios. In contrast, deep learning-based methods offer a data-driven framework for end-to-end feature extraction and adaptive fusion. Convolutional Neural Networks (CNNs)\cite{IFcnn} have been widely adopted to extract hierarchical representations from source images, capturing both low-level textures and high-level semantic information. Nonetheless, CNNs often face difficulties in modeling long-range dependencies and global contextual relationships. Transformer-based architectures\cite{ma2022swinfusion} address these limitations by employing self-attention mechanisms to capture multi-scale cross-modal interactions, thereby improving anatomical alignment and semantic consistency. And hybrid architectures that integrate CNNs and Transformers\cite{tang2022matr} have been explored to balance local feature representation and global contextual modeling, offering a more comprehensive fusion strategy. Additionally, GAN-based fusion methods\cite{ma2020ddcgan} have been proposed to improve perceptual quality through adversarial learning, but training instability remains a significant hurdle. As illustrated in fig.\ref{fig:liuchentu1}(a), most current deep learning-based fusion frameworks adopt a sequential pipeline in which the registered source images are passed through an encoder, a fusion network, and a decoder to produce the final fused result. While this approach often achieves high performance on standard visual evaluation metrics due to its reliance on learned visual features, it may still suffer from problems such as brightness stacking and the loss of fine details\cite{huang2023addns}, which can adversely affect the reliability of clinical diagnosis.

Despite significant progress made by deep learning-based fusion methods, these approaches often overlook the intrinsic semantic information inherent in medical images. Unlike natural images, where pixel intensities typically correspond to intuitive visual features, the brightness values in different medical imaging modalities convey fundamentally distinct clinical meanings\cite{fan2019semantic}. Consequently, simple channel concatenation fails to capture meaningful cross-modal correlations and may result in fused images with limited interpretability. Moreover, the evaluation of fusion performance still largely relies on conventional computer vision metrics, which inadequately reflect clinical diagnostic standards and real-world applicability. Although recent studies have attempted to incorporate textual descriptions to enhance semantic awareness in medical image fusion, their effectiveness remains limited\cite{zhao2024image}. This limitation primarily arises from the reliance on automatically generated captions by large language models such as ChatGPT. These descriptions often focus only on superficial visual attributes, such as describing "\textit{a bright circular region}" or "\textit{a darker surrounding area}", without interpreting the underlying pathological significance, for example, "\textit{a hyperintense lesion indicative of possible edema}" or "\textit{a hypodense region suggesting chronic infarction.}" In addition, they lack integration of domain-specific medical knowledge, including anatomical terminology, radiological pattern recognition, or context-aware diagnostic reasoning, which are essential for producing clinically meaningful fused outputs. As a result, the overall quality of the fusion becomes overly dependent on these shallow textual cues, significantly undermining the model’s generalizability and applicability across diverse and complex clinical scenarios.

To address the limitations of existing medical image fusion methods, this paper proposes a novel semantic-guided medical image fusion framework. Specifically, we integrate biomedical artificial intelligence models to generate descriptive texts aligned with medical images. Among these models, BiomedGPT exhibits human-level performance in generating complex radiology reports, achieving an average error rate of only 8.3\%, and has received expert-level preference ratings\cite{2023BiomedGPT}. As illustrated in Fig.~\ref{fig:liuchentu2}, we design modality-specific prompt engineering strategies to generate tailored textual prompts for different input modalities. These prompts are then fed into BiomedGPT to produce high-quality textual descriptions, which effectively guide the fusion network to attend to clinically relevant regions from multiple perspectives. This design significantly enhances both the semantic richness and clinical interpretability of the fused images. The SMFusion framework comprises three core components: text-image encoding, semantic interaction alignment, and a text injection module. Each source image is paired with its corresponding expert-level textual description and jointly embedded into a high-dimensional feature space. Semantic information is integrated into the fusion process through a text injection module based on affine transformations. Finally, BiomedGPT is employed to generate diagnostic reports from the fused images, enabling assessment of improvements in lesion depiction and localization accuracy. To ensure the preservation of fine-grained anatomical details, we introduce a medical semantic loss that encourages the fused output to retain clinically important features from the source images. Experimental results demonstrate that the proposed SMFusion framework outperforms seven state-of-the-art fusion methods in terms of both fusion quality and clinical relevance. In summary, the primary contributions of this paper can be summarized as follows:
\begin{itemize}
    \setlength{\itemsep}{0pt}  
    \setlength{\parsep}{0pt}   
    \setlength{\parskip}{0pt}  
    \item We propose a novel semantic-guided medical image fusion framework to address the low clinical utility of existing fused medical images. To the best of our knowledge, this is the first work to explicitly focus on enhancing the practical diagnostic value of fused medical images. Our framework not only generates diagnostically valuable fusion outputs but also produces corresponding diagnostic reports, introducing a new paradigm for quantifying the clinical applicability of fused images.
    \item We construct a multimodal medical image–text dataset tailored for the medical image fusion task, facilitating future research in diagnostic report generation and clinical value quantification for fused images. In this study, the dataset is designed specifically for fusion tasks involving multimodal medical images; in future work, we aim to extend it to broader multimodal medical scenarios. The dataset will be publicly released on Github soon.
    \item We introduce a semantic interaction alignment module and a medical semantic loss function to guide the generation of semantically enriched fused medical images. Our approach not only preserves key diagnostic information from the source modalities but also achieves superior performance in terms of brightness, detail preservation, and overall image quality.
    \item Experimental results demonstrate that SMFusion achieves superior performance in both qualitative and quantitative evaluations. Moreover, it is capable of generating detailed diagnostic reports from fused images, offering a novel paradigm for evaluating the practical value of medical image fusion methods.
\end{itemize}
\section{Related Work} \label{sec:Related Work}
\subsection{Multimodal Medical Image Fusion Methods}
The development of multimodal medical image fusion has progressed from handcrafted fusion rules to deep learning-based frameworks that emphasize semantic adaptability and data-driven representation learning. While traditional methods established the foundations for spatial and transform domain fusion, deep learning approaches have introduced greater flexibility in capturing the heterogeneous characteristics of medical modalities. Early deep fusion models predominantly utilized Convolutional Neural Networks (CNNs) to extract hierarchical features and perform pixel-level or feature-level fusion. Representative methods such as IFCNN\cite{IFcnn} demonstrated the ability to preserve structural details, but often failed to capture long-range dependencies and modality-specific semantics. Transformer-based architectures SwinFusion\cite{ma2022swinfusion}, have been explored to model global context via self-attention mechanisms. However, these models are typically computationally intensive and may still struggle with fine-grained spatial accuracy. Hybrid frameworks combining CNNs and Transformers\cite{tang2022matr} aim to balance local texture preservation and global semantic integration, yet their deployment is often constrained by architectural complexity and interpretability concerns. Attention mechanisms and Generative Adversarial Networks (GANs) have also been introduced to refine modality-aware feature selection and enhance perceptual realism. Nevertheless, issues such as unstable training, semantic inconsistency, and inadequate cross-modal interaction persist in many existing methods.

Although existing models have achieved significant improvements in structure preservation and perceptual fidelity, most still rely solely on visual cues and lack effective integration of high-level semantic knowledge. This limitation motivates the incorporation of external vision-language pre-trained models to enhance the interpretability and clinical relevance of fused medical images. Unlike current approaches, our method introduces textual prompts, enabling the model to focus more on regions of interest. This not only improves lesion identification in specific areas but also aligns the model’s reasoning process more closely with clinical needs.
\subsection{Generative Pre-trained Models}
In recent years, generative pre-trained models (GPMs) have emerged as a transformative paradigm in vision-language research\cite{zhou2022unsupervised}. By jointly modeling visual and textual modalities, these models achieve a unified semantic representation space, enabling cross-modal reasoning and generation. Unlike traditional models that rely on task-specific supervision, GPMs are typically trained on large-scale image-text pairs in a self-supervised or weakly supervised manner, demonstrating strong generalization across downstream tasks. Multimodal models such as CLIP\cite{clip} and BLIP\cite{blip} bridge vision and language through contrastive learning and vision-language alignment, effectively capturing high-level semantic correspondences. On the generative side, GPT-4\cite{achiam2023gpt} have demonstrated the ability to synthesize realistic images from textual prompts, extending multimodal interaction and complex scene understanding. However, despite these advances, a key limitation of large-scale models is their constrained understanding of domain-specific knowledge, particularly in specialized fields like medicine. Their generalist nature often lacks the depth required to fully grasp the intricacies of medical imagery and terminology, limiting their effectiveness in clinical applications.These challenges underscore the potential of using rich textual descriptions not only for tasks like captioning and retrieval but also for guiding fine-grained visual processes. BiomedGPT addresses this issue by integrating multimodal biomedical data to generate diagnostic-level textual descriptions from medical images\cite{2023BiomedGPT}. Its ability to process both visual and textual data makes it particularly well-suited for medical image fusion, where precise semantic alignment is crucial.

Despite the success of these models, the application of generative pre-trained models to medical image fusion remains underexplored. Most existing approaches lack explicit semantic guidance during the fusion process, limiting the clinical interpretability of the results. Our work leverages BiomedGPT to provide context-aware, diagnostic-level textual prompts, thereby enhancing the semantic fidelity and practical utility of fused medical images.
\subsection{Text Prompt Learning}
Text prompt learning has emerged as a pivotal technique in the field of vision-language models (VLMs)\cite{zhou2022learning}, leveraging textual information to guide and enhance various visual tasks. In the domain of image fusion, text prompt learning has been employed to guide the fusion process, resulting in more semantically controlled and interpretable outcomes. For example, Zhao et al. \cite{zhao2024image} introduced FILM, a novel fusion algorithm that extracts explicit textual information from source images to guide the fusion process, thereby enhancing feature extraction and contextual understanding. Additionally, Liu et al. \cite{liu2024promptfusion} proposed PromptFusion, a semantic-guided fusion approach that employs positive and negative prompts to improve the model’s ability to recognize and distinguish targets in multi-modality images. These approaches have demonstrated promising results in a range of image fusion tasks, including infrared-visible, medical, multi-exposure, and multi-focus image fusion. Integrating text prompt learning into image fusion not only improves the semantic coherence of fused images but also enhances the performance of downstream tasks such as object detection and segmentation. For instance, Cheng et al. \cite{cheng2025textfusion} introduced TextFusion, which leverages a vision-language model to establish a coarse-to-fine association mechanism between text and image signals, enabling controllable image fusion. Experimental results indicate that this approach outperforms traditional appearance-based fusion methods in terms of both visual quality and task-specific performance. Furthermore, Yi et al. \cite{yi2024text} developed Text-IF, a text-guided image fusion model that addresses issues of image degradation and non-interactivity in existing fusion methods. By leveraging semantic text guidance, this method facilitates interactive and high-quality fusion results. Overall, text prompt learning holds great potential for enhancing the performance of vision-language models and guiding image fusion tasks.
\section{Method}\label{sec:Method}
\subsection{Problem Formulation}
In most image fusion networks the input consists of two source images (e.g., $I_{1}$, $I _{2}$), and the output tends to produce a fixed final fusion result (e.g.$I^{f}$)\cite{2024Text}. This network structure $\mathcal{T}_n$, focusing on the fusion of feature dimensions can be described as:
\begin{equation}
I^{f} = \mathcal{F}(I_{1}, I_{2}; \mathcal{T}_n).
\end{equation}

Current fusion paradigms predominantly focus on visual-level representations in medical imaging. Although existing deep learning methodologies demonstrate superior performance metrics in quantitative evaluations, they exhibit significant limitations in comprehending high-level semantic information. This deficiency fundamentally conflicts with the clinical imperative to enhance the interpretability of medical imaging diagnostics. We propose an expert knowledge-based semantic-guided fusion mechanism that establishes hierarchical correlation mappings between diagnostic text descriptions and visual features. The diagnostic texts are generated by the advanced vision-language model BiomedGPT\cite{2023BiomedGPT}. This strategy not only augments the multi-dimensional information representation capacity of fused images but also maintains exceptional visual fidelity metrics.The multimodal fusion task incorporating textual prompts can be formally defined as:
\begin{equation}
    I^{f} = \mathcal{F}_{prompt}(I_{1}, I_{2}, T_{text}; \mathcal{T}_{n-t}).
\end{equation}

This text-guided controllable image fusion method integrates adjustable textual semantic prompts into the original feature mapping function $\mathcal{F}(\cdot)$, thereby constructing an enhanced mapping function $\mathcal{F}_{\text{prompt}}(\cdot)$. In particular, $\mathcal{T}_{n-t}$ denotes the image fusion network incorporating medical prior knowledge, and $T_{text}$ represents the corresponding textual descriptions used to guide the fusion process. This dynamic text-guidance mechanism enables explicit regulation of semantic spatial distribution in feature fusion, achieving multi-granularity control over the image fusion process.
\begin{figure*}[htbp] 
    \centering
    \includegraphics[width=\textwidth, trim=18 18 18 18, clip]{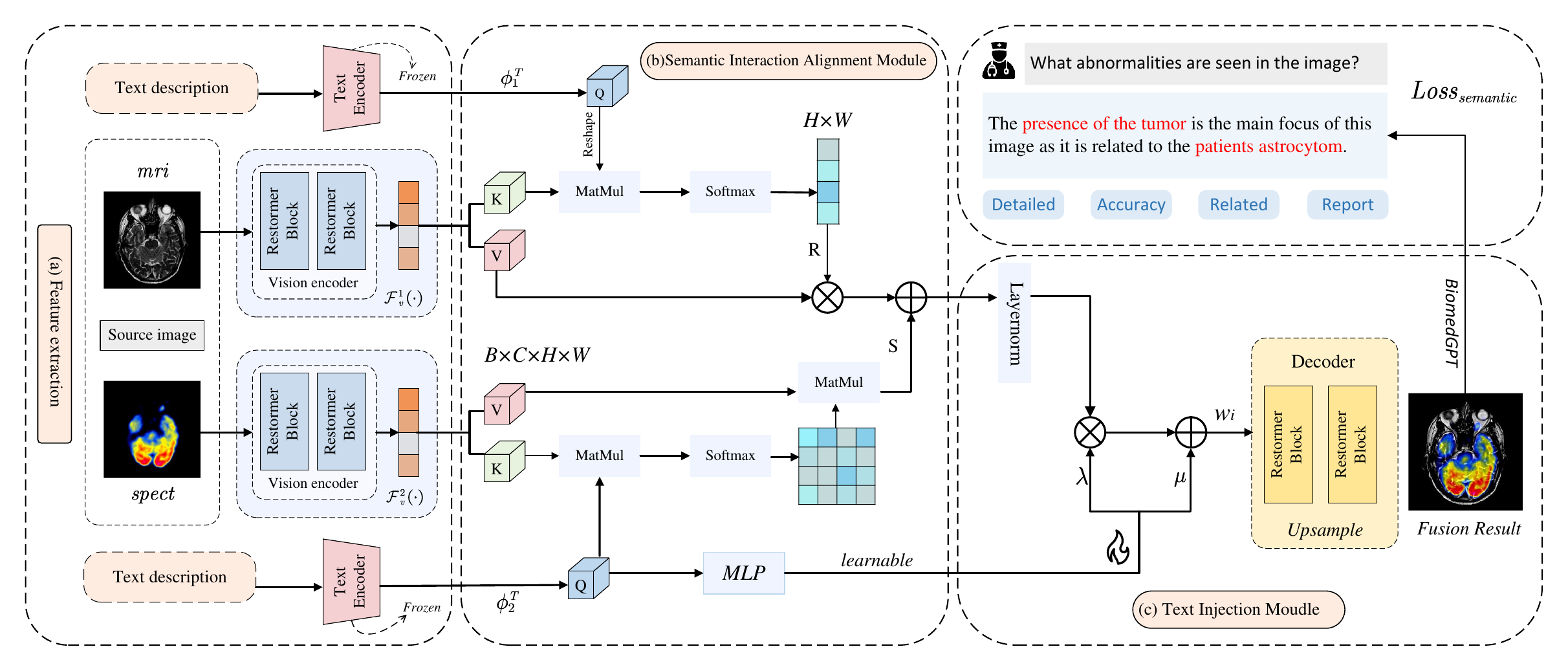} 
    \caption{The workflow of SMFusion. It consists of three key components: the feature extraction module (a), the semantic interaction alignment module (b), and the text injection module (c). The encoded text features are guided through the L-layer (b) and (c) module to facilitate image fusion.}
    \label{fig:workflow}
\end{figure*}
\subsection{Image Fusion Pipeline}
\subsubsection{Encoder} multimodal medical images contain abundant anatomical details and pathological characteristics, posing significant challenges for deep semantic extraction. Although Zhao et al.\cite{2024Text} attempted to employ ChatGPT for medical image captioning, the outputs were limited to superficial descriptions of apparent features. Building upon the biomedical AI model BiomedGPT\cite{2023BiomedGPT}, and leveraging the authoritative Harvard Medical Image Dataset, we construct professional-grade multimodal text-images dataset. The implementation workflow proceeds as follows: For input image pairs {$I_{1}, I_{2}$}, BiomedGPT generates precisely aligned pathological descriptions $\{T_1, T_2\}$. These textual outputs are then fed into a parameter-freezing CLIP text encoder to obtain semantic feature vectors $ \left\{ \phi_1^T, \phi_2^T \right\}$. Regarding image encoding, to address the spatial feature extraction requirements of multimodal medical images, we introduce dual-layer Restormer blocks to construct the feature extraction network. This architecture efficiently decouples deep semantic information from spatial topological features through the synergistic interaction of multi-head transposed attention and gated feed-forward networks\cite{zamir2022restormer}. This process can be stated as follows:\begin{equation}
    F_{1} = \mathcal{F}_v^1(I_{1}), F_{2} = \mathcal{F}_v^2(I_{2}),
\end{equation}
where $I_{1} \in \mathbb{R}^{H \times W \times 1}$ and $I_{2} \in \mathbb{R}^{H \times W \times 3}$ denote input images from two distinct modalities. $H$ and $W$ represent the height and width of the images, respectively. $\mathcal{F}_v^1$ and $\mathcal{F}_v^2$ are the encoders applied to $I_1$ and $I_2$ to extract the visual features, denoted as $F_1$ and $F_2$, from each modality.
\subsubsection{Semantic Feature Modulation}. The Semantic Feature Modulation (SFM) block is designed to achieve cross-modal dynamic regulation of fused features $F_f$ through textual feature space $\phi^T \in \mathbb{R}^{2 \times 128}$. This architecture comprises two pivotal components: a semantic interaction alignment module based on cross-attention mechanisms\cite{cheng2025textfusion,zhu2022one} and a learnable semantic injection gating module which will be introduced in Sec.\ref{sec:3.3}. To facilitate deep feature interaction and efficient semantic guidance, $L$ cascaded SFM blocks are sequentially applied through a hierarchical coupling framework, followed by the final decoding stage performed by the decoder. In simple terms, the Feature Modulation can be described as:
\begin{equation}
    F_{f}^{l+1} = \{ \mathcal{F}_{f}^{D}(\mathcal{A}_{n}^{s}(F_{f}^{l}, \phi_{i}^T)) \}_{m},
\end{equation}

where $F_f^{l}$ denotes the fused image features at the $l$-th stage. $\{\cdot\}_m$ is used to represent the hierarchical coupling architecture. $\mathcal{F}_{f}^{D}$ and $\mathcal{A}_{n}^{s}$ refer to the decoder based on Restormer Blocks and SFM, respectively. Note that the decoder includes downsampling operations corresponding to the upsampling in the encoder, ensuring consistency in the size of the feature maps.
\subsubsection{Diagnostic Report and Loss Function} The current state-of-the-art fusion frameworks employing deep neural networks often result in brightness accumulation in the output images. Despite achieving high scores in certain visual evaluation metrics, these methods inevitably lead to a loss of semantic information\cite{huang2023addns}. To address this issue, we innovatively propose a novel framework that automatically generates a diagnostic report for the fusion results. This feature is designed to assist medical professionals in reducing their workload and enhancing diagnostic efficiency. Specifically, we introduce a medical semantic loss function $\mathcal{L}(\cdot)$, which will be detailed in Sec.\ref{sec:3.4}. This loss function is designed to explicitly guide the fusion process to retain more semantic information. The final output of our model comprises two components: the fused image F and the diagnostic report.

\subsection{Semantic Guidance Fusion Architecture}\label{sec:3.3}
\subsubsection{Semantic Interaction Alignment Module} The semantic interaction alignment module enables the model to handle diverse textual prompts, thereby effectively editing the latent representations of image features and more accurately localizing lesion regions. Medical images, on one hand, typically contain rich spatial information, which is critical for precisely identifying the location of pathological areas. On the other hand, they often carry fine-grained semantic content, such as blood flow patterns that can offer higher-level contextual cues. These characteristics are complementary to the contextual information provided by textual prompts, jointly enhancing the model’s holistic understanding of medical images. Inspired by FFCLIP \cite{zhu2022one}, we establish a semantic alignment between text and image through an affine transformation, where all transformation parameters are learned via a cross-attention mechanism. As illustrated in Fig. \ref{fig:workflow}, for the output of the $i$-th semantic interaction alignment module $w_i$ (where $i \in {1, 2, \ldots, L}$ and $w_0$ is initialized as the features encoded from multimodal images), we compute cross-attention maps between $w_i$ and $\phi^T$ along both the spatial and channel dimensions, tailored to the information conveyed by each modality. For the position dimension, $\phi_{1}^T$ is passed through a convolutional block and set as the query $Q_{m} \in \mathbb{R}^{256}$, while the latent encoding $w_{i}$ is set as both the value $V \in \mathbb{R}^{256 \times 256}$ and the key $K \in \mathbb{R}^{256 \times 256}$. Subsequently, we calculate the cross-attention map to obtain the scaling parameter $R$. This parameter undergoes matrix multiplication with the value on the position dimension, so as to transform and match the source image feature space with its corresponding text semantic embedding. Briefly, This process can be expressed as:
\begin{equation}
\begin{array}{l}
Q_{m}=\phi_{1}^T W^{Q}, K=w_{i} W^{K}, V=w_{i} W^{V}, \\
R = {Softmax}\left(\frac{Q_{m}K^{T}}{d_k}\right), \\
\end{array}
\end{equation}
where $W^{Q},W^{K},W^{v} \in \mathbb{R}^{256 \times 256}$, $d_{k}$ is the scaling factor, we are able to dynamically adjust the contribution degree of different dimensions via these parameters, given the semantic prompt $\phi^T_{1}$. As the features along the positional dimension alone are insufficient to capture the complete semantic attributes, it remains necessary to compute the parameters along the channel dimension of the source image, thereby further enhancing semantic alignment.

For the channel dimension, similar to the position dimension, we utilize the same K and V as those for the position dimension, along with a new query to compute the feature map within the channel dimension, then, the V is reconstructed based on this feature map. Therefore, it can be described as:
\begin{equation}
\begin{array}{l}
Q_{n}=\phi_{2}^T W^{Q}_{n}, K=w_{i} W^{K}, V=w_{i} W^{V}, \\
S = {Softmax}\left(\frac{Q_{n}K}{d_k}\right), \\
\end{array}
\end{equation}
where $W^{Q}_{n} \in \mathbb{R}^{1 \times 256}$,  After obtaining the scale and translation parameters, we can obtain the semantically aligned features\cite{zhu2022one} according to the following predefined linear transformation rules:
\begin{equation}
    \hat{F}_f^i = R \times V + S,
\end{equation}
where $\hat{F}_f^i$ represents the output of the entire semantic feature alignment module. Subsequently, by passing through the text injection module, we are able to acquire the fused image features that incorporate semantic information. This feature space has the potential to enhance the representational capacity of the image.

\subsubsection{Text Injection Module} In the text injection module, the aligned fused feature space $\hat{F}_f^i$ is combined using the semantic learning parameters $\lambda$ and $\mu$, which guide feature fusion from the perspectives of scale adjustment and bias control, respectively, so as to obtain the final fused feature $F_f$. Therefore, it is defined as:
\begin{equation}
    F_f = (1+\lambda) \frac{\hat{F}_f^i-\mu_{{\hat{F}_f^i}}}{\sigma_{\hat{F}_f^i}}+\mu,
\end{equation}
 where $\mu_{{\hat{F}f^i}}$ and $\sigma_{\hat{F}_f^i}$ represent the mean and variance of the feature that incorporates text semantic information, respectively, which can help achieve more stable and efficient convergence during model training. The semantically learned parameters $\lambda$ and $\mu$ are obtained via a multi-layer perceptron (MLP), which consists of two fully connected layers with the ReLU activation function.

\subsection{Loss Functions}\label{sec:3.4}
The loss function plays a crucial role in guiding the network optimization process, directing it to optimize in the correct direction and largely determining the quality of the fused image. From the perspective of semantic-guided image fusion, we aim for the fused image to not only preserve clear texture details and luminance information but also exhibit a high degree of similarity to the given text prompts. In this section, we introduce two main components of the loss function: Visual Quality Loss, which includes gradient loss and reconstruction loss, and Semantic Consistency Loss, which ensures that the fused image aligns well with the semantic content provided by the text prompts.

\subsubsection{Medical semantic loss} We designed a medical semantic loss function based on \cite{wei2022hairclip}, which quantifies the semantic correlation between the fused image and the text prompt. This loss function operates at the semantic level, ensuring that the generated image aligns closely with the textual description. By measuring the discrepancy between the image and the text in a shared semantic space, this loss function directly enhances the model's ability to preserve the intended meaning conveyed by the text.  It is defined as follows:
\begin{equation}
\mathcal{L}_{\text{semantic}}=\left\{ \begin{array}{l}
	0\,\,,\ \ \ \ if\,\,\,\,\cos \left( \mathcal{F}_{v}^{i}\left( I^f \right) ,\phi ^T \right) \geq \theta\\
	1-\cos \left( \mathcal{F}_{v}^{i}\left( I^f \right) ,\phi ^T \right) ,otherwise\\
\end{array} \right. ,
\end{equation}
where $F_{v}^i$ denotes the encoder for the input source images, $\phi ^T$ represents the encoded textual features. $\theta$ is a hyperparameter. Through extensive experiments, we set it to 0.85 as the text serves merely as a prompt, which only needs to guide the fused image to optimize towards semantic similarity at the initial stage.

\subsubsection{Gradient consistency} Gradient Consistency: Once the text similarity loss exceeds the threshold $\theta$, the gradient will no longer be updated. Therefore, to maximize the preservation of the texture and useful information of the source images and further enhance the quality of the fused image, we introduce the gradient loss\cite{liu2024semantic}. It can be expressed as follows:
\begin{equation}
    \mathcal{L}_{g r a d}=\frac{1}{H W}\left\|\nabla F_{f}-\max \left(\nabla F_{1}, \nabla F_{2}\right)\right\|_{1},
\end{equation}
where $\nabla$ denotes the Sobel gradient operator, and $\max(\cdot)$ represents the operation of selecting the maximum value for each element, with $H$ and $W$ denoting the height and width of the image, respectively.

\subsubsection{Reconstruction loss} The reconstruction loss is primarily employed to guarantee the information integrity between the source image and the fused image throughout the entire encoding process\cite{liu2024promptfusion}, ensuring that the regions of interest in the semantic prompts can be effectively preserved. It can be specifically expressed as follows:
\begin{equation}
    \mathcal{L}_{1} = ||F_{1} - F_{f}|| + \omega \mathcal{L}_{SSIM}(\mathbf{F}_{1}, \mathbf{F}_{f}),
\end{equation}
where $L_{ssim} = 1 - SSIM$, $SSIM(\cdot)$ represents the structural similarity index\cite{zhao2016loss}, and $\omega$ is the adjustment factor. Similarly, $\mathcal{L}_{2}$ can also be obtained using the same method.

\subsubsection{Total loss} The total loss is a linear combination of the relevant losses mentioned above, and it is formulated as follows:
\begin{equation}
\mathcal{L}_{total}=\alpha\mathcal{L}_{semantic}+\beta\mathcal{L}_{grad}+\gamma\mathcal{L}_{rec},
\end{equation}
where $\alpha$, $\beta$, and $\gamma$ are adjustable hyper-parameters associated with the fusion task, which play a decisive role in determining the quality of the final fused image.
\section{Experiments}\label{sec:Experiments}
\subsection{Implementation Details and Datasets}
\subsubsection{Implementation Details} The training and testing datasets for proposed SMFusion were both obtained from \textit{Harvard} medical school and can be publicly available at \href{http://www.med.harvard.edu/AANLIB/home.html}{http://www.med.harvard.edu/AANLIB/home.html}. The text descriptions were generated using BiomedGPT fine-tuned with instruction, and the images had a size of 256×256. Specifically, the dataset comprised 357 MRI-SPECT pairs, 269 MRI-PET pairs, and 184 MRI-CT pairs. During the training process, the learning rate was set to 0.0001, and the Adam optimizer was employed to train the network for 150 epochs with a batch-size of 2. Furthermore, the hyperparameter $\alpha$ was cautiously set to 0.1, while $\beta$ and $\gamma$ were determined to be 1.5 respectively. In the Retormer block, the dimension of each attention head was precisely set to 64. All the experiments mentioned above were completed on a Tesla T4 GPU and an Intel Core i5-12500H CPU with the PyTorch framework.
\subsubsection{Comparison Approaches} We compared the proposed fusion method SMFusion with seven state-of-the-art methods, namely DIFF-IF\cite{Diff-if}, MetaFusion\cite{Metafusion}, EMMA\cite{EMMA}, DDFM\cite{DDFM}, IFCNN\cite{IFcnn}, U2Fusion\cite{u2fusion}, and SDNet\cite{sdnet}. The codes of all these comparison methods are publicly available. Moreover, we strictly set these methods according to the parameter configurations given in the corresponding papers to ensure the scientificity and rigor of the comparison.
\subsubsection{Evaluation Metrics} To quantitatively evaluate the fusion results, we employ five widely-used metrics: spatial frequency (SF)\cite{SF}, average gradient (AG)\cite{AG}, standard deviation (SD)\cite{SD}, edge information transfer ($Q_{a b / f}$)\cite{huang2023addns}, and multi-scale structural similarity index measure (MS-SSIM)\cite{wang2003multiscale}. These metrics capture various critical aspects of fusion quality, including texture structure, edge sharpness, and overall visual fidelity, providing a comprehensive assessment of the method's performance in medical image fusion tasks.

$Q_{ab/f}$  can quantify the edge information transfer between the original images and the fused image, thereby reflecting the quality of visual information in the fused image. The higher the value of $Q_{ab/f}$, the better the quality of the fused image. It is written as follows:
\begin{equation}
Q_{a b / f}=\frac{\sum\limits_{i=1}^{H} \sum\limits_{j=1}^{W} \left( Q_{a f}(i, j) w_{1}(i, j) + Q_{b f}(i, j) w_{2}(i, j) \right)}{\sum\limits_{i=1}^{H} \sum\limits_{j=1}^{W} \left( w_{1}(i, j) + w_{2}(i, j) \right)},
\end{equation}
where $H$ and $W$ denote the height and width of the source images respectively, and $Q_{a f}(i,j)$ is define as:
\begin{equation}
    Q_{f m}(i, j)=Q_{f m}^{e}(i, j) Q_{f m}^{g}(i, j),
\end{equation}
where $Q_{f m}^{e}$ and $Q_{f m}^{g}$ indicate the edge intensity and orientation retention measures at position $(i,j)$, $Q_{bf}(i, j)$ follows the same definition logic. $\omega_{1}$ and $\omega_{2}$ denote the weights derived from the gradient intensity of the source image, respectively.

SD is a measure that captures the distribution and contrast of the fused image. It indicates the variability of pixel values relative to the mean, and images with higher contrast exhibit superior visual representation effects. It can be expressed as follows:
\begin{equation}
SD = \sqrt{\frac{1}{MN} \sum\limits_{i=1}^{M} \sum\limits_{j=1}^{N} (x_{i,j} - \mu)^2},
\end{equation}
where  $x_{i,j}$ denote the pixel value and $\mu$ signifies the average pixel value.

SF represents the overall activity within an image in the spatial domain. It is defined by the following equation:
\begin{equation}
    SF=\sqrt{RF^2+CF^2} ,
\end{equation}
where RF and CF denote the row frequency and column frequency. 
\begin{figure*}[htbp] 
    \centering
    \includegraphics[width=\textwidth,trim=18 18 18 18, clip]{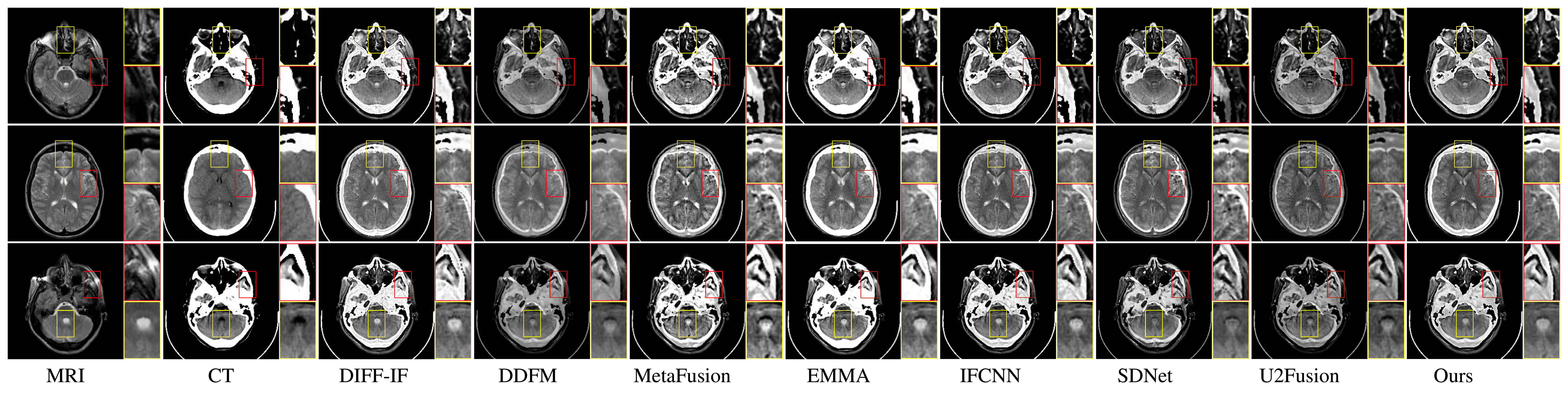} 
    \caption{Qualitative comparison of our SMFusion with seven state-of-the-art methods on three MRI and CT image pairs. Enlarged detail patches are highlighted with red and yellow boxes for visualization.}
    \label{fig:MRI-CT-Result}
\end{figure*}
AG evaluates an image's ability to preserve details and textures and is commonly used to assess image sharpness. For a given image of size M×N, the average gradient is formulated as:
\begin{equation*}
    \bar{G}=\frac{\sum_{i=1}^{M-1}\sum_{j=1}^{N-1}\frac{1}{4}\sqrt{\left(\frac{\partial g(i,j)}{\partial i}\right)^2+\left(\frac{\partial g(i,j)}{\partial j}\right)^2}}{(M-1)(N-1)},
\end{equation*}
where \( (m, n) \) represents the coordinates of the image, and $\frac{\partial g}{\partial i}$ and $\frac{\partial g}{\partial n}$ denote the horizontal and vertical gradient components, respectively. A higher average gradient value indicates a richer representation of image details, signifying a better-fused result.

MS-SSIM can represent the multi-scale assessment of image quality, considering the human visual system's sensitivity to structural information at different scales. The overall MS-SSIM index is calculated by combining the luminance, contrast, and structure comparisons at each scale, weighted by their respective importance.  It can be formulated as:
\begin{equation}
    MS(x, y) = \left[ l_M(x, y) \prod_{j=1}^M [c_j(x, y)]^{\alpha_j} [s_j(x, y)]^{\beta_j} \right]^{\gamma_j},
\end{equation}
where \( M \) is the number of scales, and \( \alpha_j \), \( \beta_j \), and \( \gamma_j \) are parameters that define the relative importance of each scale.
\subsection{Comparison with SOTA methods}
\subsubsection{Qualitative analysis} To further validate the effectiveness of SMFusion as a semantic-guided image fusion paradigm, we selected brain images from three representative patients for fusion experiments. As shown in Fig.\ref{fig:MRI-CT-Result}, compared with seven state-of-the-art fusion methods, SMFusion exhibits three distinct advantages. SMFusion accurately delineates the location and size of the lesion, providing more reliable diagnostic information. It maintains a high level of color fidelity and hierarchical information during the fusion process, ensuring effective integration of multimodal information. Finally, high-density structures such as bones in CT images do not completely obscure the soft tissue details in MRI images, thereby achieving a balanced information representation. Specifically, the fusion results generated by DDFM, SDNet, and U2Fusion exhibit low brightness, leading to significant structural detail loss and reducing visual quality. While MetaFusion produces images with higher brightness, it introduces substantial artifacts, limiting its fusion performance. EMMA excessively incorporates information from CT images, resulting in an inaccurate depiction of the lesion region. In contrast, DIFF-IF and IFCNN effectively capture complementary information from the original images; however, they exhibit limitations in preserving fine soft tissue details from MRI images. Overall, SMFusion outperforms existing methods in lesion representation, color fidelity, and balanced information preservation.

\begin{figure*}[htbp] 
    \centering
    \includegraphics[width=\textwidth,trim=18 18 18 18, clip]{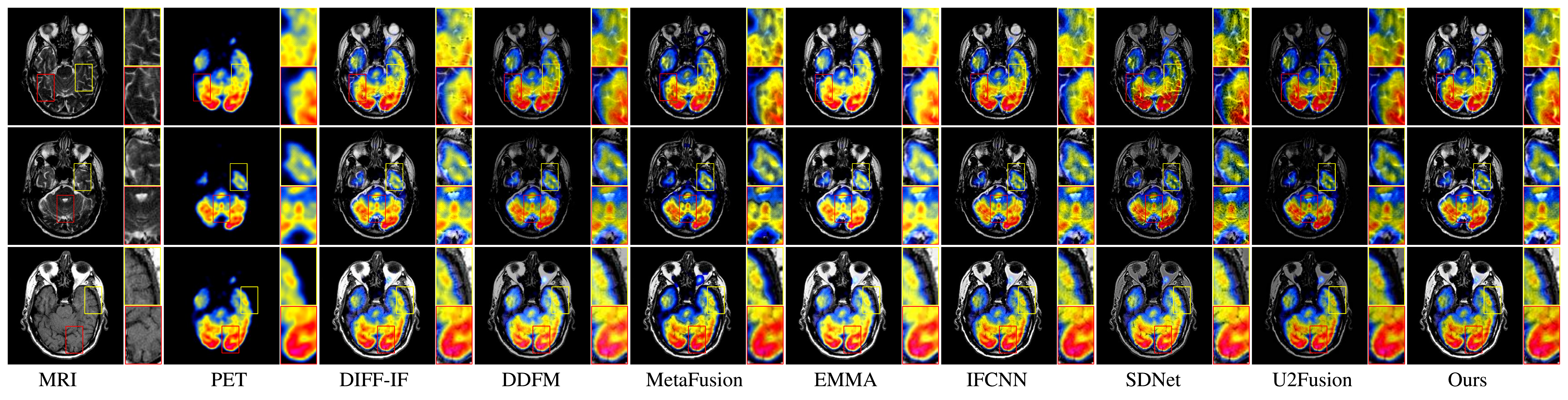} 
    \caption{Qualitative comparison of SMFusion with seven state-of-the-art methods on three MRI and PET image pairs. Notably, our method demonstrates superior performance in blood flow representation and texture preservation (e.g., the first row).}
    \label{fig:MRI-PET-Result}
\end{figure*}

\begin{figure*}[htbp] 
    \centering
    \includegraphics[width=\textwidth,trim=18 15 18 18 ,clip]{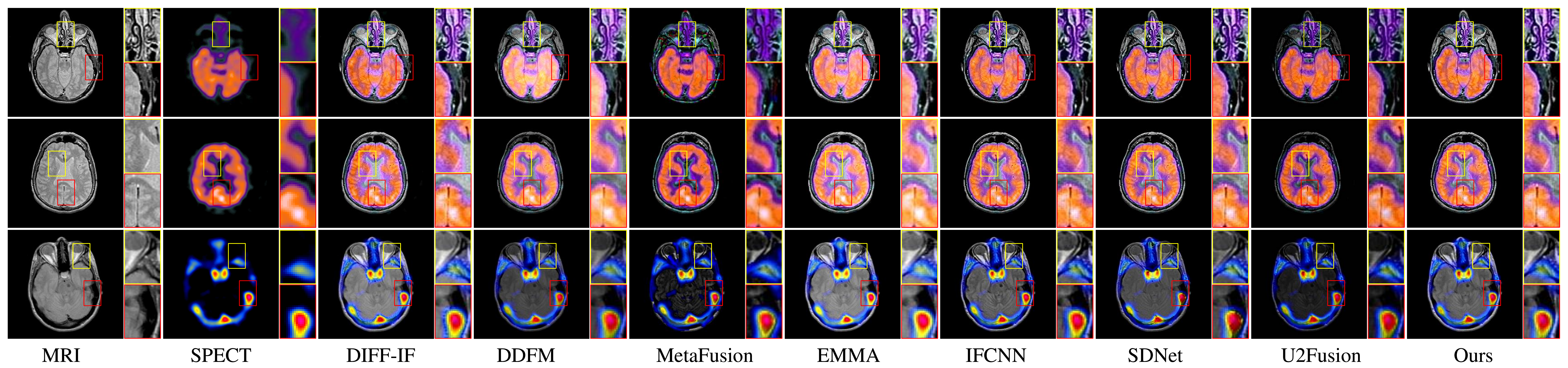} 
    \caption{Qualitative comparison of ours with seven state-of-the-art methods on three MRI and SPECT image pairs. Enlarged detail patches are highlighted with red and yellow boxes and displayed on the right side of the original images for visualization.}
    \label{fig:MRI-SPECT-Result}
\end{figure*}

As shown in Fig.\ref{fig:MRI-PET-Result}, we conducted a subjective qualitative analysis of MRI-PET fusion images. DDFM and EMMA exhibit strong color extraction capabilities for PET images. However, their unbalanced information extraction strategy leads to brightness accumulation, which negatively impacts the preservation of cortical folds in MRI images. In contrast, SDNet and U2Fusion show lower performance in information extraction and color consistency, failing to fully capture the essential features from the original images. MetaFusion suffers from noticeable color distortion, particularly evident in the first set of fused images. Additionally, while DIFF-IF and IFCNN perform well in preserving edge and texture details, they fall short in integrating soft tissue structures and functional information compared to SMFusion. This limitation is particularly observable in the first set of fusion results. The MRI-SPECT fusion results are shown in Fig.\ref{fig:MRI-SPECT-Result}. Similar to the MRI-PET fusion results, SMFusion effectively extracts true color information while preserving rich texture details. In the first row of fusion results, MetaFusion and U2Fusion fail to effectively extract soft tissue information from MRI images and exhibit significant color deviations from the source images, which degrades the overall fusion quality. In contrast, SMFusion excels in integrating structural information from MRI and functional information from SPECT, ensuring that essential features are preserved while maintaining high visual quality. Compared to other methods, SMFusion achieves a well-balanced fusion of multimodal information, avoiding information loss or excessive enhancement, and delivering superior visual coherence. In general, SMFusion outperforms the other seven state-of-the-art methods in both visual quality and structural detail preservation, further validating its effectiveness as a semantic-guided image fusion paradigm.
\begin{table*}[ht]
    \centering
    \renewcommand{\arraystretch}{1}
    \caption{Quantitative comparison of our SMFusion with seven state-of-the-art methods on the test dataset in terms of mean and standard deviation. Red and blue fonts indicate the best and second-best results, respectively.}
    \label{tab:comparison}
    \resizebox{0.8\textwidth}{!}{%
    \begin{tabular}{l|l l l l l} 
        \toprule
         Method & \textbf{SF$\raisebox{-0.9ex}{\scalebox{0.9}{$\uparrow$}}$} & \textbf{AG$\raisebox{-0.9ex}{\scalebox{0.9}{$\uparrow$}}$} & \textbf{MS\_SSIM$\raisebox{-0.9ex}{\scalebox{0.9}{$\uparrow$}}$} & \textbf{SD$\raisebox{-0.9ex}{\scalebox{0.9}{$\uparrow$}}$} & \(\mathbf{Q}_{a b / f}\) \raisebox{-0.9ex}{\scalebox{0.9}{$\uparrow$}}\\
        \midrule
        \textbf{MetaFusion}  & 17.742 \,$\pm$\, 7.839  & 4.734 \,$\pm$\, 2.347  & 1.160 \,$\pm$\, 0.337  & 46.457 \,$\pm$\, 2.347  & 0.453 \,$\pm$\, 3.302  \\
        \textbf{DDFM}       & 16.317 \,$\pm$\, 4.636  & 4.189 \,$\pm$\, 1.328  & \textcolor{blue}{1.298 \,$\pm$\, 0.185}  & 56.287 \,$\pm$\, 10.827  & 0.635 \,$\pm$\, 0.101  \\
        \textbf{IFCNN}      & 23.135 \,$\pm$\, 7.74   & 5.852 \,$\pm$\, 2.048  & 1.297 \,$\pm$\, 0.201  & 55.066 \,$\pm$\, 10.641  & 0.702 \,$\pm$\, 0.058  \\
        \textbf{U2Fusion}   & 12.738 \,$\pm$\, 3.944  & 3.346 \,$\pm$\, 1.216  & 1.129 \,$\pm$\, 0.241  & 33.733 \,$\pm$\, 10.58   & 0.378 \,$\pm$\, 0.062  \\
        \textbf{SDNet}      & 21.958 \,$\pm$\, 6.241  & 5.501 \,$\pm$\, 1.779  & 1.235 \,$\pm$\, 0.199  & 49.137 \,$\pm$\, 8.759   & 0.647 \,$\pm$\, 0.071  \\
        \textbf{EMMA}       & 21.809 \,$\pm$\, 6.576  & 5.601 \,$\pm$\, 1.951  & 1.263 \,$\pm$\, 0.204  & \textcolor{red}{69.128 \,$\pm$\, 10.701}  & 0.592 \,$\pm$\, 0.089  \\
        \textbf{DIFF-IF}    & \textcolor{blue}{23.852 \,$\pm$\, 7.428}  & \textcolor{blue}{6.065 \,$\pm$\, 1.978}  & 1.276 \,$\pm$\, 0.207  & 59.410 \,$\pm$\, 10.246  & \textcolor{red}{0.756 \,$\pm$\, 0.058}  \\
        \textbf{Ours}       & \textcolor{red}{25.770 \,$\pm$\, 8.034}  & \textcolor{red}{6.333 \,$\pm$\, 2.055}  & \textcolor{red}{1.298 \,$\pm$\, 0.197}  & \textcolor{blue}{60.230 \,$\pm$\, 10.674}  & \textcolor{blue}{0.707 \,$\pm$\, 0.064}  \\
        \bottomrule
    \end{tabular}%
    }
\end{table*}
\subsubsection{Quantitative analysis} To objectively evaluate the fusion performance of SMFusion, we randomly selected 50 brain fusion images from different patients and quantitatively compared them with seven state-of-the-art fusion algorithms. The quantitative results are presented in Table.\ref{tab:comparison}. Experimental results demonstrate that SMFusion achieves the highest performance in SF, AG and MS-SSIM, indicating that our method effectively preserves more details and texture information from the original images while avoiding brightness accumulation issues, resulting in the clearest fusion images. These findings are highly consistent with the qualitative evaluation results. For SD and Qabf metrics, SMFusion achieves the second-best performance. This is primarily due to the introduction of a semantic interaction alignment module in our method, which enhances the fused image’s ability to understand textual descriptions, thereby extracting richer semantic features from the source images. Unlike other methods, our approach prioritizes the retention of high-level semantic information, which may result in a slight compromise in these specific quantitative metrics. However, this trade-off ensures that the fused images effectively preserve critical medical details, such as lesion regions, making them more clinically valuable. Overall, these results confirm that SMFusion achieves a well-balanced fusion outcome, maintaining both high perceptual quality and essential diagnostic information.
\subsection{Performance in Generating Diagnostic Reports}
\subsubsection{Comparative Analysis of Different Methods} 
Medical image fusion aims to provide physicians with more accurate and comprehensive diagnostic information. However, generating diagnostic reports from fused images remains labor-intensive and time-consuming. Existing research primarily focuses on improving the visual quality of fused images, with limited attention given to their effectiveness in downstream clinical applications. To address this gap, we randomly selected 27 brain images from the test set and leveraged the medical reasoning capabilities of BiomedGPT, which has demonstrated human-level performance in generating complex radiology reports, to generate diagnostic reports based on different fusion methods. This model's ability to synthesize detailed, context-aware reports makes it particularly suited for clinical use. We evaluated eight fusion approaches quantitatively across three dimensions: average text length, information entropy, and keyword count. These dimensions provide a balanced assessment of the reports, capturing aspects such as depth, informativeness, and relevance, while reflecting both the richness and clarity of the diagnostic information. The evaluation results are summarized in Table \ref{Diagnose report} and Table \ref{tab:fusion_table}, where we present a comparison with SDnet due to space constraints. Additional comparison results can be found in the appendix.

Experimental results demonstrate that our proposed method achieves the highest performance in terms of information entropy, indicating that the generated diagnostic reports contain more semantically rich medical information. As shown in Table \ref{tab:fusion_table}, a direct comparison between our method and SDNet, the best-performing baseline, further validates this advantage. In the first set of fusion results, both methods successfully identified a defect in the posterior fossa. However, our method additionally detected an ependymoma and indicated potential pathological abnormalities. In the second set, when BiomedGPT was applied to SDNet’s fusion output, it merely highlighted regions of increased metabolic activity without providing meaningful medical insights.

In addition, we also show the diagnostic reports of single-modal and multi-modal multi-channel (non-fusion) medical images in Table \ref{Report 2}. It is evident that these pre-fusion images exhibit certain limitations in conveying comprehensive diagnostic information. Specifically, both single-modality and non-fused multi-modality images are only capable of capturing partial brain features, such as cerebral edema in the right cerebellum, morphological changes in the sella turcica, and enlargement of the pituitary gland, or merely referencing brain metabolic activity. Such information has some diagnostic value; however, it does not offer a systematic or comprehensive view. For example, in the second case presented, neither the single-modality image nor the combined information from two separate modalities was sufficient to accurately identify the lesion. In contrast, the fused image clearly reveals the presence of an adenoma and provides interpretable lesion characteristics, so as to assist the doctor to further confirm the accuracy of the results.
\begin{figure}[htbp] 
    \centering
    \includegraphics[width=\linewidth,trim=18 18 18 18 ,clip]{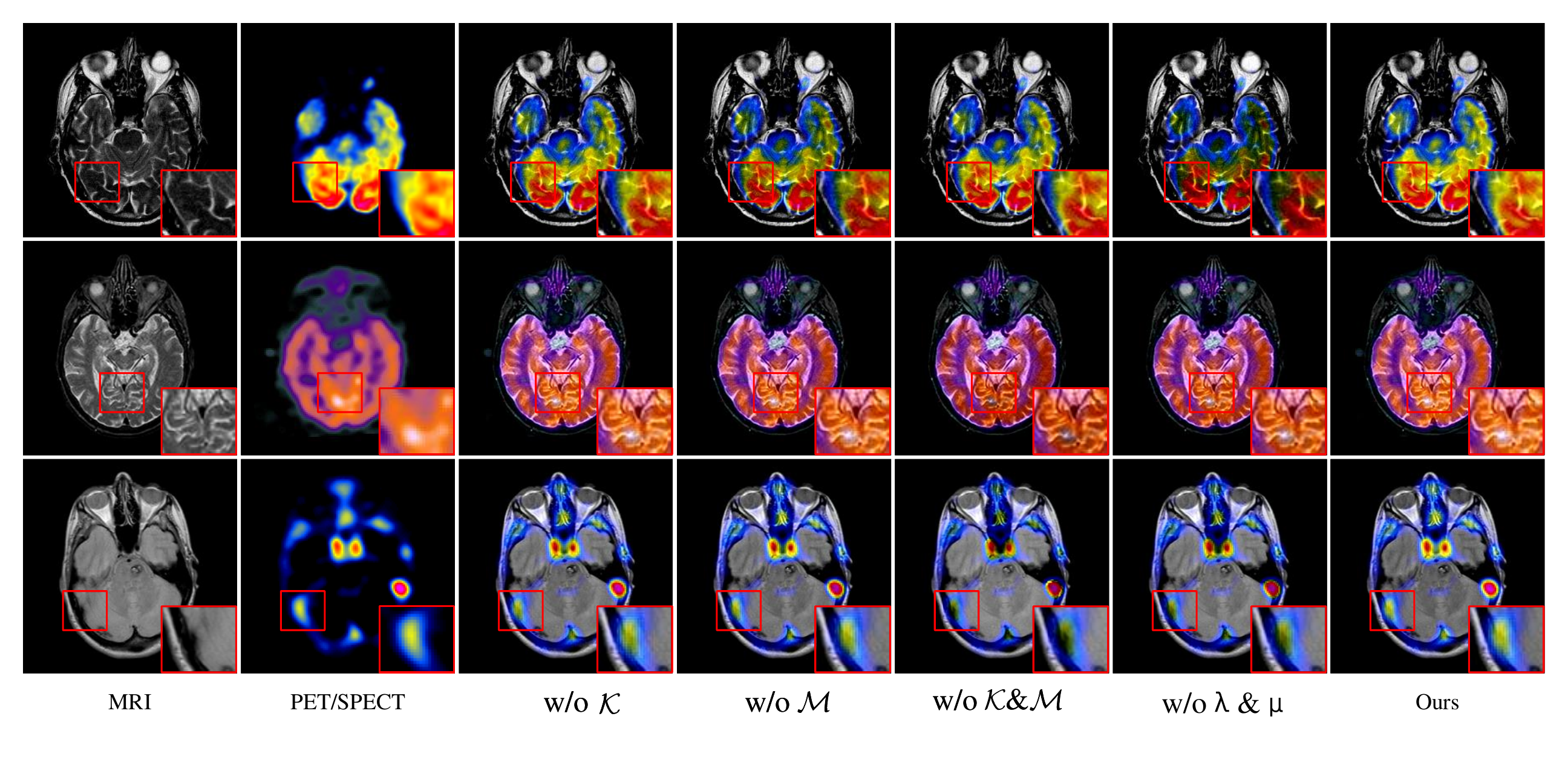} 
    \caption{Ablation qualitative experiments were performed on three different image pairs. Each module shows a positive contribution to the fusion performance of SMFusion.}
    \label{Ablation studies Fig1}
\end{figure}
\begin{figure}[htbp] 
    \centering
    \includegraphics[width=0.9935\linewidth,trim= 18 18 18 18 ,clip]{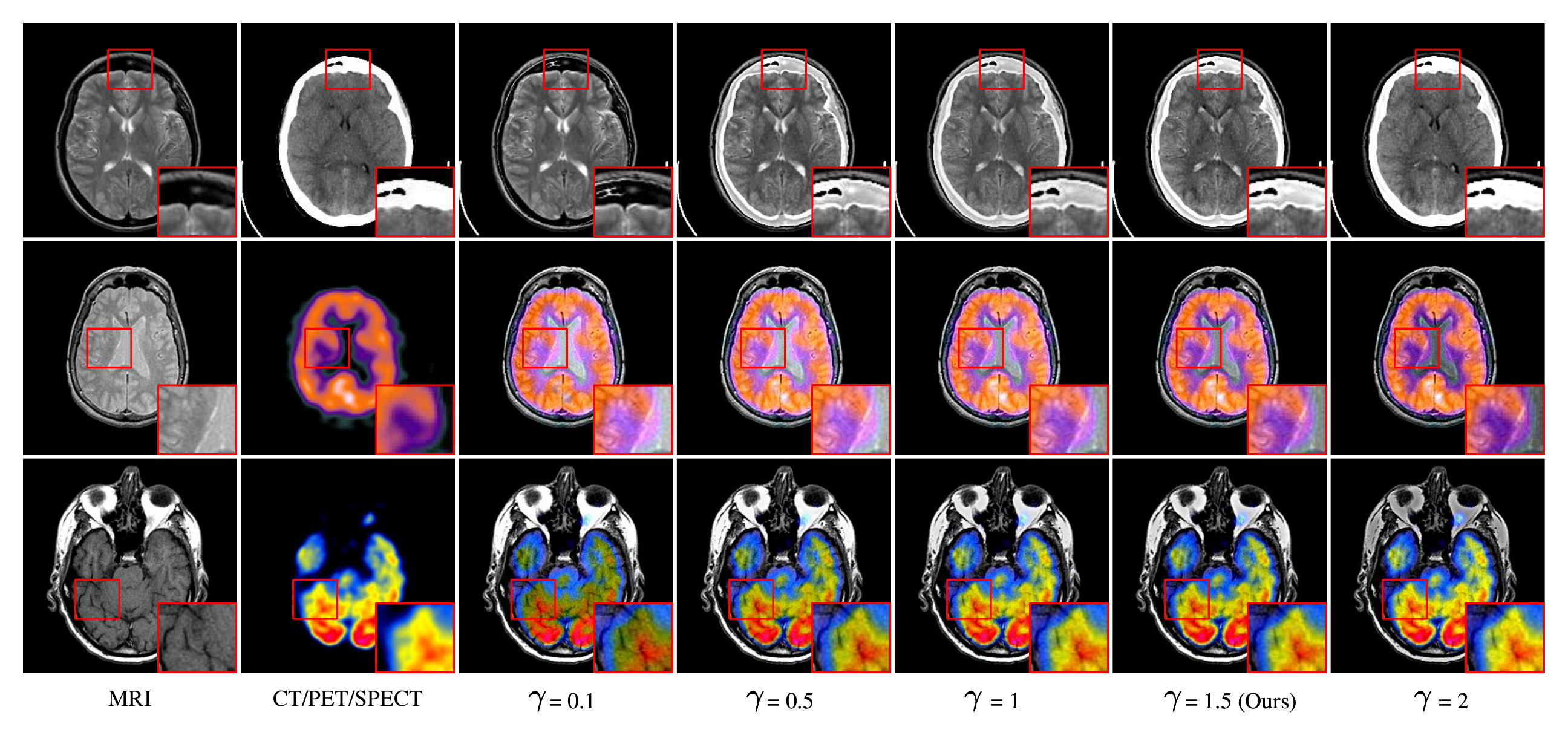} 
    \caption{Ablation experiments between different hyperparameter settings. The enlarged details are displayed in the bottom-right corner of the source image.}
    \label{Ablation studies Fig2}
\end{figure}
\subsubsection{Mean Opinion Score testing}
To ensure the reliability of the diagnostic reports generated by large language models, we conducted a questionnaire-based subjective assessment. Specifically, we recruited 13 medical students with medical backgrounds to participate in the study. Each participant was asked to rate the diagnostic reports using one of three levels: Moderate, Satisfactory, or Excellent, based on how well the content aligned with clinical expectations. We randomly selected 10 diagnostic reports based on fused images from the test set, covering a total of 8 experimental groups representing 8 different fusion methods. To eliminate potential bias, all method-related information was blinded to the participants. The results of the survey are summarized in Table \ref{wjdc}. As shown, over 80\% of the reports across all fusion methods received a rating of Satisfactory or above. Notably, the reports generated based on our fusion results received the highest proportion of Excellent ratings. These findings suggest that the majority of diagnostic reports generated by BiomedGPT based on fused images are considered accurate.
\subsection{Ablation studies}
To validate the effectiveness of the proposed method, we conducted a systematic analysis of the key innovations in SMFusion, focusing on the impact of the semantic interaction alignment module, loss function and the influence of different hyperparameter settings in the loss function on model performance.
\subsubsection{Ablation analysis of different modules} In this paper, we propose a deep network incorporating a semantic interaction alignment module and apply it to medical image fusion. To simplify the analysis, we decompose the network into two main components, $i.e.$, $\mathcal{K}$ and $\mathcal{M}$, which represent the semantic alignment modules for different modalities. Furthermore, the semantic learning parameters $\lambda$ and $\mu$ in the text injection module facilitate the fusion of aligned features in a high-dimensional space. As shown in Fig.\ref{Ablation studies Fig1}, qualitative results demonstrate that each module contributes positively to the final fusion image. Specifically, the absence of $\mathcal{K}$ leads to suboptimal preservation of fine details and edge information, resulting in less effective enhancement of soft tissue structures in MRI images. Meanwhile, omitting $\mathcal{M}$ results in a decline in color processing capability. Furthermore, when both $\mathcal{K}$ and $\mathcal{M}$ are removed, the network fails to learn sufficient medical semantic information, making it unable to focus on abnormal regions in the source images, as illustrated in the second set of fusion results.
\begin{figure}[htbp]
    \centering
    \includegraphics[width=\linewidth, trim=18 18 18 18, clip]{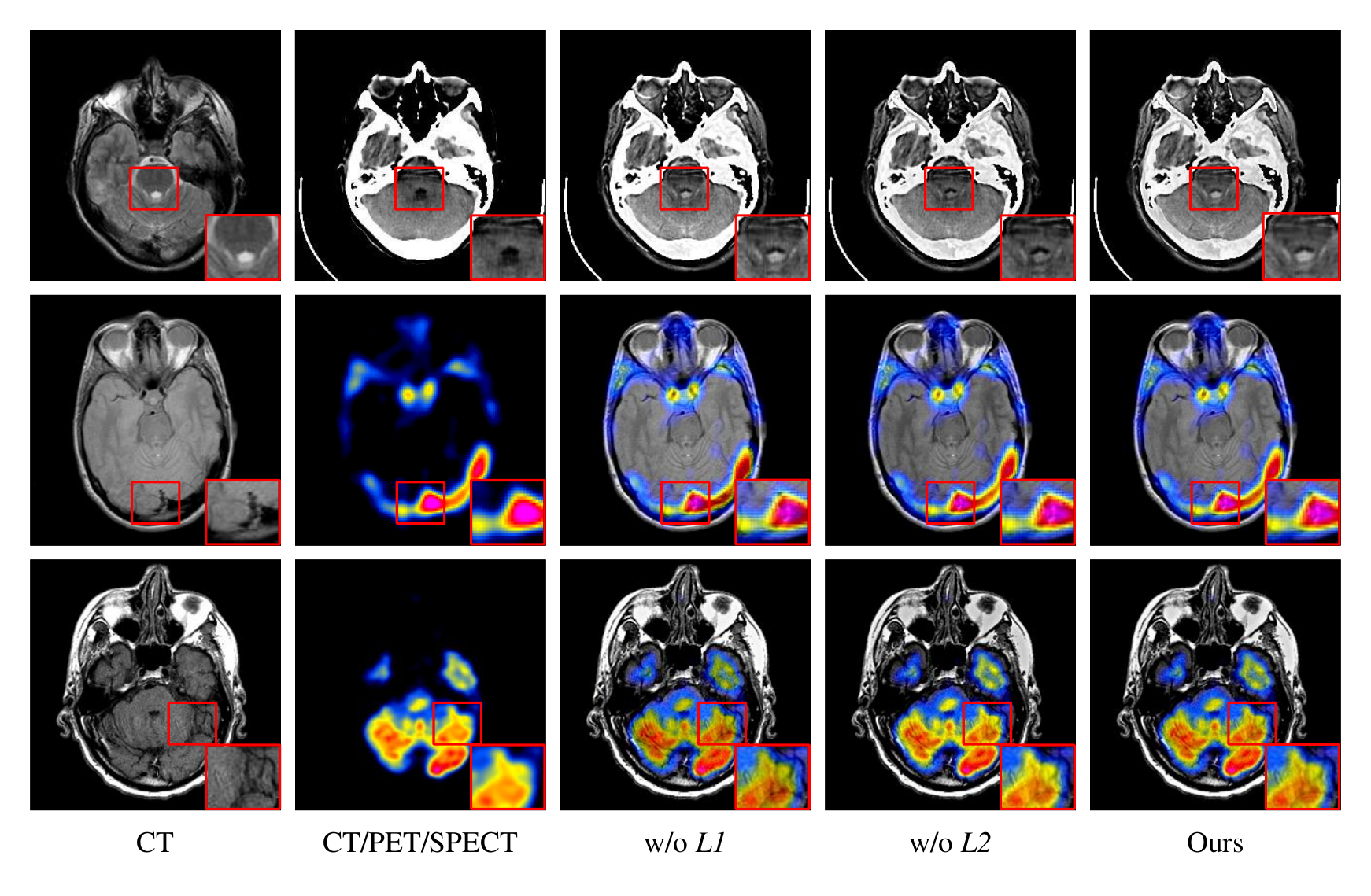}
    \caption{Qualitative analysis of different loss functions. $L1$ and $L2$ represent the proposed medical semantic loss and gradient loss, respectively.}
    \label{Ablation studies Fig3}
\end{figure}
\begin{table}[ht]
    \centering
    \caption{Quantitative analysis under different combinations of loss functions. Red and blue indicate the best and second-best results, respectively. The fusion results are optimal when all proposed loss functions are incorporated.}
    \renewcommand{\arraystretch}{1}
    \label{Ablation studies table3}
    \setlength{\tabcolsep}{10pt}
    \resizebox{\linewidth}{!}{%
    \begin{tabular}{l|lllll}
        \toprule
        Metric & \textbf{SF} & \textbf{AG} & \textbf{MS-SSIM} & \textbf{SD} &  $\textbf{Q}_{ab/f}$ \\
        \midrule
        w/o $l1$ & $25.355$ & $6.333$ & $1.286$ & $59.258$ & $0.701$ \\
        w/o $l2$ & \textcolor{red}{$25.829$} & $6.316$ & $1.293$ & $59.646$ & $0.700$ \\
        ours & \textcolor{blue}{$25.770$} & \textcolor{red}{$6.335$} & \textcolor{red}{$1.298$} & \textcolor{red}{$60.230$} & \textcolor{red}{$0.707$} \\
        \bottomrule
    \end{tabular}}
\end{table}
\begin{table*}[ht]
    \centering
    \renewcommand{\arraystretch}{1} 
    \setlength{\tabcolsep}{10pt} 
    \caption{Quantitative comparison of the fusion results of different methods in generating diagnostic reports. Red and blue fonts indicate the optimal and second-best results, respectively.}
    \label{Diagnose report}
    \resizebox{\textwidth}{!}{ 
    \begin{tabular}{l|llllllll} 
        \toprule
        Method & \textbf{DDFM} & \textbf{Diff-IF} & \textbf{EMMA} & \textbf{IFCNN} & \textbf{MetaFusion} & \textbf{U2Fusion} & \textbf{SDNet} & \textbf{Ours} \\
        \midrule
        \textbf{Text Length} & 40.2593 & \textcolor{blue}{41.8148} & 37.6296 & \textcolor{red}{43.0370} & 39.5185 & 38.9259 & 40.0370 & 33.8889 \\
        \textbf{Entropy} & 4.0490 & 4.0633 & 4.0250 & 4.0604 & 4.0561 & 4.0490 & \textcolor{blue}{4.0707} & \textcolor{red}{4.0762} \\
        \bottomrule
    \end{tabular}
    }
\end{table*}
\begin{table*}[ht]
    \centering
    \renewcommand{\arraystretch}{1.3} 
    \setlength{\tabcolsep}{2pt}       
    \caption{Diagnostic report generation from fusion images produced by SDNet and SMFusion. Red text indicates key diagnostic information. Clearly, SMFusion generates more detailed and specific descriptions.}
    \label{tab:fusion_table}
    \resizebox{0.95\textwidth}{!}{ 
    \begin{tabular}{@{} >{\centering\arraybackslash}m{2cm} 
                        >{\centering\arraybackslash}m{2.1cm} 
                        >{\raggedright\arraybackslash}m{6cm} 
                        >{\centering\arraybackslash}m{1.6cm} 
                        >{\centering\arraybackslash}m{1.6cm} 
                        >{\raggedright\arraybackslash}m{2.6cm} @{} }
        \toprule
        \textbf{Method} & \textbf{Image} & \multicolumn{1}{c}{\textbf{Fusion Result Description}} & \textbf{Text Length} & \textbf{Entropy} & \textbf{Key Words} \\
        \midrule
        Ours  
        & \includegraphics[width=1.5cm]{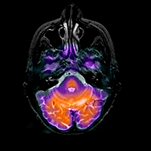}  
        & Defect in the \textcolor{red}{posterior fossa} which is an abnormality in the optic nerve sheath, a rare congenital condition where there are \textcolor{red}{ependymomas} found, as well as an area of increased radiotracer uptake on the surface of the brain, indicating a possible \textcolor{red}{abnormality or pathology}. 
        & 44 & 4.0130 & posterior fossa, ependymomas, abnormality or pathology \\
        
        SDNet  
        & \includegraphics[width=1.5cm]{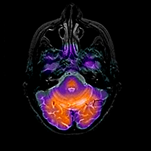}  
        & Defect in the \textcolor{red}{posterior fossa} which is an abnormality in the brain tissue as seen on the other side of the cerebellum, midbrain, and pons, as well as the presence of a \textcolor{red}{lacunar infarct}, a type of brain tissue that stands out from the surrounding healthy brain tissue. 
        & 48 & 3.9535 & posterior fossa, lacunar infarct \\
        
        Ours  
        & \includegraphics[width=1.5cm]{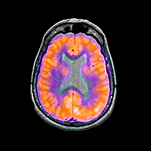}  
        & Brain structures such as the basal ganglia and lateral ventricles are also visible in the image, but the main focus is on the \textcolor{red}{adenoma}, a type of brain tumor with increased uptake of the radiotracer due to the \textcolor{red}{presence of a lacunar infarct}. 
        & 43 & 3.9739 & adenoma, presence of a lacunar infarct \\
        
        SDNet  
        & \includegraphics[width=1.5cm]{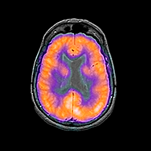}  
        & Positron emission tomography is a technique used to assess the metabolic activity of the brain. In this image, you can see areas of \textcolor{red}{increased metabolic activity}, represented by brighter colors compared to the surrounding brain tissue. 
        & 36 &  4.0725 & increased metabolic activity \\
        
        \bottomrule
    \end{tabular}
    }
\end{table*}
\begin{table*}[ht]
    \centering
    \renewcommand{\arraystretch}{1} 
    \setlength{\tabcolsep}{10pt} 
    \caption{Quantitative ablation of different modules, with red and blue marking the best and second-best results.}
    \label{Ablation studies table1}
    
    \resizebox{\textwidth}{!}{%
    \begin{tabular}{cccc|lllll} 
        \toprule
        $\mathcal{K}$ & $\mathcal{M}$ & $\lambda$ & $\mu$ & SF & AG & MS\_SSIM & SD & $\textbf{Q}_{ab/f}$ \\
        \midrule
        \checkmark & \checkmark &  &  & 25.679 $\pm$ 8.025 & 6.308 $\pm$ 2.111 & 1.210 $\pm$ 0.213 & 60.018 $\pm$ 10.523 & 0.675 $\pm$ 0.075 \\
          &  & \checkmark & \checkmark & 25.691 $\pm$ 8.019 & 6.304 $\pm$ 2.060 & 1.276 $\pm$ 0.202 & 57.711 $\pm$ 11.001 & 0.706 $\pm$ 0.064 \\
          & \checkmark & \checkmark & \checkmark & 25.671 $\pm$ 8.128 & \textcolor{blue}{6.323 $\pm$ 2.088} & \textcolor{blue}{1.278 $\pm$ 0.203} & 59.783 $\pm$ 10.955 & 0.697 $\pm$ 0.062 \\
        \checkmark &  & \checkmark & \checkmark & \textcolor{red}{26.182 $\pm$ 8.327} & 6.315 $\pm$ 2.149 & 1.266 $\pm$ 0.204 & \textcolor{blue}{60.210 $\pm$ 11.282} & \textcolor{blue}{0.703 $\pm$ 0.063} \\
        \checkmark & \checkmark & \checkmark & \checkmark & \textcolor{blue}{25.770 $\pm$ 8.034} & \textcolor{red}{6.335 $\pm$ 2.055} & \textcolor{red}{1.298 $\pm$ 0.197} & \textcolor{red}{60.230 $\pm$ 10.674} & \textcolor{red}{0.707 $\pm$ 0.064} \\
        \bottomrule
    \end{tabular}%
    }
\end{table*}
Further analysis reveals that removing the text injection module prevents the high-dimensional text-image feature space from being fully disentangled, introducing additional noise during decoding and causing noticeable color distortions in the fused images. This indicates that the text injection module plays a crucial role in extracting effective image information and significantly enhances the network's robustness. Moreover, as shown in table \ref{Ablation studies table1}, quantitative experimental results further validate the proposed method. Our approach achieves state-of-the-art performance across six evaluation metrics. Each module contributes to different aspects of fusion quality improvement, demonstrating the positive impact of incorporating both the semantic interaction alignment module and the text injection module on model performance.
\begin{table*}[ht]
    \centering
    \renewcommand{\arraystretch}{1} 
    \setlength{\tabcolsep}{10pt} 
    \caption{Quantitative analysis of the ablation experiments with different hyperparameters, where the fusion performance achieves the best results across all metrics when $\gamma$ is set to 1.5.}
    \label{Ablation studies table2}
    \resizebox{\textwidth}{!}{%
    \begin{tabular}{l|lllll} 
        \toprule
        The value of \textbf{$\gamma$} & \textbf{SF} & \textbf{AG} & \textbf{MS\_SSIM} & \textbf{SD} & $\textbf{Q}_{ab/f}$ \\
        \midrule
        $\gamma = 0.1$ & $24.699 \pm 6.721$ & $6.066 \pm 2.118$ & $1.159 \pm 0.172$ & $53.196 \pm 14.843$ & $0.668 \pm 0.083$ \\
        $\gamma = 0.5$ & $24.525 \pm 7.661$ & $6.281 \pm 2.017$ & $1.265 \pm 0.202$ & $60.025 \pm 10.599$ & \textcolor{red}{$0.734 \pm 0.055$} \\
        $\gamma = 1$ & $24.356 \pm 7.970$ & \textcolor{red}{$6.478 \pm 2.104$} & $1.272 \pm 0.206$ & \textcolor{blue}{$60.122 \pm 11.133$} & $0.700 \pm 0.066$ \\
        $\gamma = 2$ & \textcolor{blue}{$25.149 \pm 7.982$} & $6.136 \pm 1.959$ & \textcolor{blue}{$1.280 \pm 0.207$} & $55.606 \pm 12.473$ & $0.680 \pm 0.061$ \\
        $\gamma = 1.5$ (Ours) & \textcolor{red}{$25.770 \pm 8.034$} & \textcolor{blue}{$6.335 \pm 2.055$} & \textcolor{red}{$1.298 \pm 0.197$} & \textcolor{red}{$60.230 \pm 10.674$} & \textcolor{blue}{$0.707 \pm 0.064$} \\
        \bottomrule
    \end{tabular}%
    }
\end{table*}
\begin{table*}[!t]
    \centering
    \renewcommand{\arraystretch}{1} 
    \setlength{\tabcolsep}{10pt} 
    \caption{Results of the subjective assessment, where the percentages represent the proportion of participants' ratings for diagnostic reports. Notably, fewer moderate ratings and more other ratings are better.}
    \label{wjdc}
    \resizebox{\textwidth}{!}{ 
    \begin{tabular}{l|cccccccc} 
        \toprule
        Method & \textbf{DDFM} & \textbf{Diff-IF} & \textbf{EMMA} & \textbf{IFCNN} & \textbf{MetaFusion} & \textbf{U2Fusion} & \textbf{SDNet} & \textbf{Ours} \\
        \midrule
        Moderate & \textcolor{red}{7\%} & 16\% & 19\% & 12\% & 19\% & 16\% & 18\% & \textcolor{blue}{9\%} \\
        Satisfactory & \textcolor{red}{66\%} & \textcolor{blue}{59\%} & 52\% & 62\% & 47\% & 51\% & 54\% & 56\% \\
        Excellent & 27\% & 25\% & 29\% & 26\% & 34\% & \textcolor{blue}{33\%} & 28\% & \textcolor{red}{35\%} \\
        \bottomrule
    \end{tabular}
    }
\end{table*}
\begin{table}[!t]
    \centering
    \renewcommand{\arraystretch}{0.9}  
    \setlength{\tabcolsep}{4pt}         
    \caption{Diagnostic Reports for Single-Modality and Multi-Modality Multi-Channel (Non-Fusion) Scanned Images.}
    \label{Report 2}
    \begin{tabular}{@{}>{\centering\arraybackslash}m{2.5cm} 
                    >{\raggedright\arraybackslash}m{0.65\linewidth}@{}}
        \toprule
        \textbf{Image} & \textbf{Diagnostic Report} \\
        \midrule
        \includegraphics[width=1.35cm]{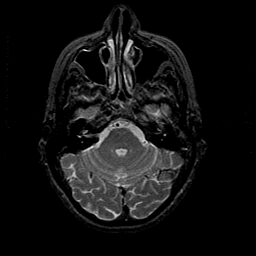}  & Brain edema can be seen in the image as an area of increased signal intensity within the right cerebellum, which is located between the brainstem and the cerebrum. \\
        \includegraphics[width=1.35cm]{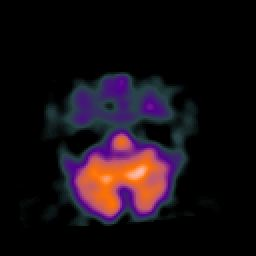} & There is a noticeable reduction in the size of the sella turcica and an enlargement of the pituitary gland, which is the main focus of the scan. \\
        \includegraphics[width=1.35cm]{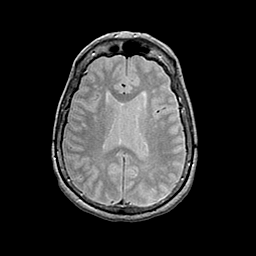}  & Remnant Astrocytoma, Brain Edema, Brain Nonenhancing Tumor. \\
        \includegraphics[width=1.35cm]{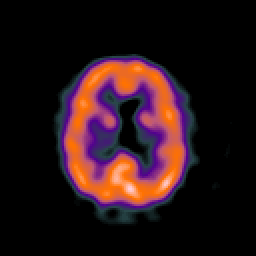} & Emission computed tomography scans provide detailed information about the brain’s metabolic activity, such as inflammation, dysplastic changes, or pathological changes. \\
        \bottomrule
    \end{tabular}
\end{table}
\subsubsection{Ablation analysis of different hyperparameter} Extensive experiments have demonstrated that the choice of hyper-parameters plays a crucial role in determining the quality of the fusion results. The loss function updates gradients through backpropagation, guiding the network toward minimizing the gradient and ultimately influencing the quality of the fused images. In our experiments, we observed that the hyperparameter $\gamma$ in the reconstruction loss significantly affects the model’s fusion performance. The qualitative analysis results, as shown in Fig.\ref{Ablation studies Fig2}, indicate that we set $\gamma$ to 0.1, 0.5, 1, 1.5, and 2 to examine the variation in fusion images. It was observed that as the value of $\gamma$ increased, more CT information was retained in the fused images, which led to the suppression of soft tissue structures from the MRI images, as shown in the first group of example. The quantitative analysis results, presented in Table \ref{Ablation studies table2}, further confirm this trend, demonstrating that our method achieves superior performance across most metrics and remains in a leading position. Considering both qualitative and quantitative evaluations, the optimal fusion results are obtained when $\gamma$ is set to 1.5, as it ensures a balanced extraction of information from the source images.
\subsubsection{Ablation analysis of loss function} In Section \ref{sec:3.4}, we introduce a medical semantic loss function specifically designed to enhance fusion performance in medical imaging tasks. The qualitative fusion results under different loss combinations are shown in Fig.~\ref{Ablation studies Fig3}. As evident from the visual comparisons, the absence of the $L1$ loss leads to undesirable artifacts and noticeable color distortions in the fused images. When the $L2$ loss is excluded during training, the fusion results tend to miss critical structural details, such as the continuity of soft tissue in MRI images. The quantitative results summarized in Table \ref{Ablation studies table3}. Our method achieves the best performance across all evaluated metrics, demonstrating the effectiveness and necessity of the proposed loss functions in improving fusion quality.
\section{Conclusion} \label{sec:Conclusion}
In this study, we propose a semantic-guided medical image fusion method that incorporates expert-level prior knowledge, allowing the model to focus more precisely on pathological regions while preserving critical medical information. This approach enhances the visibility of lesions and ensures a well-balanced integration of fine details and high-level semantic information. To further improve fusion performance, we introduce a medical semantic loss, aligning the fused images more closely with real-world clinical needs in terms of content consistency and medical relevance. In addition, we develop a novel evaluation mechanism that assesses the effectiveness of fused images through the diagnostic reports they generate, providing a clinically oriented perspective on fusion quality. Experimental results demonstrate that our method outperforms current approaches in preserving fine details, enhancing texture representation, and integrating semantic information, offering valuable support for clinical decision-making. In the future, we will further investigate the mechanisms of text-guided multimodal medical image fusion and conduct extensive validation on large-scale clinical datasets to improve model robustness and generalization.
\bibliographystyle{IEEEtran}  
\bibliography{Reference}     
\end{document}